# Hierarchical Maximum Margin Learning for Multi-Class Classification

**Jian-Bo Yang**
School of Computer Engineering
Nanyang Technological University
Singapore, 639798

**Ivor W. Tsang**
School of Computer Engineering
Nanyang Technological University
Singapore, 639798

## Abstract

Due to myriads of classes, designing accurate and efficient classifiers becomes very challenging for multi-class classification. Recent research has shown that class structure learning can greatly facilitate multi-class learning. In this paper, we propose a novel method to learn the class structure for multi-class classification problems. The class structure is assumed to be a binary hierarchical tree. To learn such a tree, we propose a maximum separating margin method to determine the child nodes of any internal node. The proposed method ensures that two class-groups represented by any two sibling nodes are most separable. In the experiments, we evaluate the accuracy and efficiency of the proposed method over other multi-class classification methods on real world large-scale problems. The results show that the proposed method outperforms benchmark methods in terms of accuracy for most datasets and performs comparably with other class structure learning methods in terms of efficiency for all datasets.

## 1 Introduction

Multi-class classification is a common pattern recognition problem. Traditionally, it is often solved by decomposing the multi-class problem into several two-class problems. Suppose multi-class problem has $c$ classes. Conventional methods like one-versus-one (1vs1) and one-versus-rest (1vsR) decompose the $c$-class problem into $c(c-1)/2$ and $c$ binary classification problems respectively. However, practitioners often encounter with problems having hundreds or thousands of classes. These problems are rather challenging for conventional methods. Particularly, in the case where instantaneous prediction is required, 1vs1 and 1vsR methods become computationally expensive since $c(c-1)/2$ or $c$ classifiers need to be evaluated for the prediction of each instance's label.

To alleviate such computational burdens, class structure learning has been proposed in the past several years (Platt et al., 2000; Beygelzimer et al., 2009a,b; Bengio et al., 2010). Basically, class structure is depicted by a graph that characterizes the relations among all decomposed classification problems. For simplicity, class structure can be assumed as a binary tree, and its definition is given by:

**Definition 1.1** *A tree of c-class structure has d layers. Its root node is $\{1, 2, \cdots, c\}$. At layer $t$, $\forall t = 1, 2, \cdots, d$, there are $n_t$ node(s) where $n_t \geq 1$ ($n_1 = 1$) and each node $G_i^t \subseteq \{1, 2, \cdots, c\}, \forall i = 1, 2, \cdots, n_t$ is a group of classes. This tree is constrained by: 1) each non-leaf node has two children and each child node is a subset of its parent node; 2) the nodes in the same layer $t$, $\forall t > 1$, have non-overlapping class indices, i.e., $G_i^t \cap G_j^t = \emptyset$ for $i \neq j$ where $1 \leq i, j \leq n_t$.*

When the class structure tree is constructed, the label of an instance is predicted by traversing the tree along the path from the root node to a leaf node. This cost, on average, is sublinear with respect to $c$, i.e. $O(\log(c))$. Obviously, it has computational advantage over 1vs1 and 1vsR methods. Note that class structure learning is not only for facilitating multi-class learning but also useful in many applications, e.g. protein structure learning (Murzin et al., 1995), image structure prediction (Berg et al., 2010) and information retrieval (Manning et al., 2008).

The key problem of class structure learning is how to determine the internal nodes in the tree. It is still quite challenging to obtain all internal nodes at one time, so researchers often resort to some iterative schemes to obtain the nodes layer by layer, either from top to bottom or from bottom to top. The scheme from top to bottom is adopted in the proposed method, due to its simplicity and efficiency. In this case, tree construction problem becomes a set of internal node splitting problems. Typically, if an internal node contains $c^*$ classes, there are $2^{(c^*-1)} - 1$ possible ways to generate its children. Among them, which one is the best?

To answer this question, we first propose a criterion, Separating Margin criterion, to evaluate the goodness of a pair of child nodes for an internal node. Herein, the separating margin is the minimum distance between the boundary of data in two class-groups that are represented by the two child nodes. With this criterion, we then propose a MSM model, short for Maximum Separating Margin, to determine the optimal pair of child nodes. The proposed model is a mixed integer programming (MIP) problem which is known to be computationally expensive in general. To solve this MIP problem, an approach using convex relaxation is employed to reformulate this MIP problem as an easily-solved convex problem w.r.t. a continuous variable. This reformulated problem can be cast into the multiple kernel learning (MKL) framework. Consequently, existing MKL solvers can be used to solve the proposed model. Moreover, the cutting plane method is also used to expedite the convergence of the MKL procedure.

When MSM model is repeatedly used for all internal nodes in the tree from the root to the leaves, the class structure tree is constructed. The resultant tree ensures that two class-groups represented by any two sibling nodes have the maximum separating margin. In experiments, the proposed method in comparison with benchmark methods are tested on large scale multi-class datasets. The results show that the proposed method outperforms benchmark methods in terms of accuracy for most datasets and performs comparably with other class structure learning methods in terms of efficiency for all datasets.

The rest of this paper is organized as follows. Section 2 reviews some related multi-class classification methods with/without class structure learning. Section 3 presents the proposed method. Section 4 discusses an extensive set of experiments for performance evaluation over large-scale datasets. Section 5 concludes this paper.

## 2 Related Work

This section reviews the existing multi-class classification methods with and without class structure learning. Some of them will serve as benchmarks in numerical experiments.

We begin with the notations used in this paper. Dataset $\mathcal{D} = \{x_i, y_i\}_{i=1}^N$ is assumed to be given with $x_i \in \mathcal{X}$ as $i^{th}$ instance and $y_i \in \{1, \cdots, c\}$ as the corresponding label. Symbols **0** and **1** are the vectors with all zeros and all ones respectively. $I_m$ is the $m \times m$ identity matrix. $[x]_i$ is the $i$-th element of $x$. $|S|$ is the cardinality of the set $S$. $A \odot B$ is the element-wise production of vectors $A$ and $B$.

### 2.1 Classification without Class Structure Learning

Multi-class classification is often solved by the combinatorial methods without class structure learning. Besides the well-known 1vs1 and 1vsR as mentioned before, error-correcting and "all-together" methods are also widely used. Error-correcting method (Escalera et al., 2010) is based on 1vs1 and 1vsR but modifies them with various outputs coding and decoding schemes. "All-together" methods developed in (Weston and Watkins, 1999; Crammer et al., 2001; Keerthi et al., 2008) directly consider all 1vsR classifiers together in one formulation for training, and it uses the same 1vsR procedure in testing. The comparison of all above methods can be found in (Rifkin and Klautau, 2004; Hsu and Lin, 2002; Keerthi et al., 2008). For all these methods, the cost of predicting the label of an instance is among the range of $O(c)$ to $O(c(c-1)/2)$, which are high for large $c$.

### 2.2 Classification with Class Structure Learning

Multi-class classification with structure learning has drawn many attentions recently. Platt et al. (2000) proposed a decision directed acyclic graph (DDAG) method. This method has the same training process with 1vs1 method but uses a different testing procedure, in which the so-called directed acyclic graph is heuristically constructed. This acyclic graph has $c(c-1)/2$ internal nodes and $c$ leaves, and each internal node is one of the 1vs1 classifiers. The cost of this method for predicting the label of an instance is $O(c)$.

Recently, some methods using the similar class structure in Definition 1.1 have been proposed, e.g., filter tree (FT) (Beygelzimer et al., 2009b), conditional probability tree (CPT) (Beygelzimer et al., 2009a) and the method proposed by Bengio, Weston and Grangier (BWG) (Bengio et al., 2010). Specifically, FT randomly chooses a binary tree as the class structure. CPT uses an online learning method to build the tree. In this method, each node is additionally associated with a decision function. All training instances sequentially traverse previously determined nodes. If the label of the coming instance have not been seen by a node, this node is split into two child nodes according to some heuristic criteria; otherwise the decision function of this node is updated by considering the new added instance. This procedure is repeated until all training data have traversed the tree, and then the resultant tree is taken as the class structure. The method BWG learns the class structure by two steps: 1), create the confusion matrix from the results of 1vsR method; 2), use graph cut algorithm for the confusion matrix to partition each internal node into several child nodes. The second step is repeatedly conducted till all leaf nodes are reached. The comparisons of above three methods' performance are given in (Bengio et al., 2010; Beygelzimer et al., 2009a). Since these three methods use the similar tree framework in Definition 1.1, they can also achieve the sublinear testing cost $O(\log(c))$ for predicting the label of an instance.

## 3 The Proposed Method

In this section, we present the proposed method for one internal node splitting problem first. Then, its generalization of the overall tree construction is given subsequently.

### 3.1 Maximum Separating Margin Model

Considering any non-leaf node $G$ containing $c^*$ classes, $\omega_1, \cdots, \omega_{c^*}$, as given in Definition 1.1, its two child nodes $G_1$ and $G_2$ are determined by the following proposed MSM0 model.

**MSM0:**

$$(G_1, G_2) = \arg\max_{\tilde{G}_1, \tilde{G}_2} \left\{ J(\tilde{G}_1, \tilde{G}_2) \mid \text{all possible } (\tilde{G}_1, \tilde{G}_2) \right\} \quad (1)$$

where $J(\tilde{G}_1, \tilde{G}_2) = \frac{2}{\|w\|^2}$ with $w$ solved by

$$\min_{w, \boldsymbol{\xi} \geq \mathbf{0}} \frac{1}{2}\|w\|^2 + \frac{C}{2}\sum_{i:y_i \in \tilde{G}_1} \xi_i^2 + \frac{C}{2}\sum_{j:y_j \in \tilde{G}_2} \xi_j^2$$
$$\text{s.t. } w'\phi(x_i) \geq 1 - \xi_i, \; \forall i : y_i \in \tilde{G}_1 \quad (2)$$
$$\quad - w'\phi(x_j) \geq 1 - \xi_j, \; \forall j : y_j \in \tilde{G}_2.$$

In the above formulas, $J(\tilde{G}_1, \tilde{G}_2)$ is the separating margin between two class-groups represented by $\tilde{G}_1$ and $\tilde{G}_2$, $\phi(x_j)$ is the vector in the high dimensional Hilbert space, $\mathcal{H}$, mapped by the function $\phi(\cdot) : \mathcal{X} \mapsto \mathcal{H}$, $w$ is the normal vector of hyperplane $\{\phi(x)|w'\phi(x) = 0\}$ in $\mathcal{H}$, $C > 0$ is a regularization parameter, $\boldsymbol{\xi} = \{\xi_j | j \in \mathcal{I}\}$ is the set of slack variables for constraints in (2).

In model MSM0, the determined pair of $(G_1, G_2)$ in (1) is the pair with the maximum separating margin, and the separating margins of all possible pairs are computed by (2) that is a standard support vector machine (SVM) model. The motivation of model MSM0 is clear: among $2^{(c^*-1)} - 1$ possible pairs of child nodes, the larger the separating margin the better the pair of child nodes. This implies that the determined two class-groups represented by $G_1$ and $G_2$ are mutually most separable. This motivation, to some extent, complies with the rule of thumb addressed in (Bengio et al., 2010) that group together classes into the same class group that are likely to be confused.

It would be computationally expensive to solve problem MSM0, as it needs to solve SVM problem (2) for $2^{(c^*-1)} - 1$ times. Next, model MSM1 is proposed to approximate MSM0 by introducing a new label variable $z \in \{+1, -1\}$ to each instance $x_j$, $\forall j \in \mathcal{I} \equiv \{i | y_i \in G\}$.

**MSM1:**

$$\min_{\boldsymbol{z} \in \mathcal{Z}} \min_{w, \boldsymbol{\xi}} \frac{1}{2}\|w\|^2 + \frac{C}{2}\sum_{j \in \mathcal{I}} \xi_j^2 \quad (3)$$
$$\text{s.t. } z_j w'\phi(x_j) \geq 1 - \xi_j, \forall j \in \mathcal{I}$$

and any class $\omega_k \in G$, $k = 1, \cdots, c^*$, is put in $G_1$ or $G_2$ by

$$\begin{cases} \omega_k \to G_1, & \text{if } z_j = 1 \\ \omega_k \to G_2, & \text{if } z_j = -1 \end{cases}, \forall j \in \{i | y_i = \omega_k\}. \quad (4)$$

In the min-min optimization problem (3), $\boldsymbol{z} = \{z_j | j \in \mathcal{I}\}$ with $z_j \in \{\pm 1\}$ and

$$\mathcal{Z} = \left\{ \boldsymbol{z} \; \middle| \; \begin{array}{l} -\beta \leq \mathbf{1}'\boldsymbol{z} \leq \beta, \text{ where } \beta \geq 0, \\ z_j = z_i \text{ if } y_j = y_i, \; i \neq j, \; \forall i, j \in \mathcal{I} \end{array} \right\} \quad (5)$$

is domain of $\boldsymbol{z}$ and $\beta$ is the user-specified parameter to balance the size of two child nodes.

**Remark 1** *The first constraint in (5) is to prevent the trivial solution $\boldsymbol{z} = \{z_j | j \in \mathcal{I}\} = \mathbf{1}$ or $-\mathbf{1}$, while the second constraint is to enforce that all instances in the same class have the same label variable $z$.*

The idea of MSM1 is that we use variable $\boldsymbol{z}$ to indicate the class-group to which each instance should belong. As shown in (4), once $\boldsymbol{z}$ is solved, the node $G$ is split into $G_1$ and $G_2$.

Since SVM problem is often solved by its dual form, so model (3) can be rewritten as

$$\min_{\boldsymbol{z} \in \mathcal{Z}} \max_{\boldsymbol{\alpha} \in \mathcal{A}} \; -\frac{1}{2}\boldsymbol{\alpha}'\left(K \odot (\boldsymbol{z}\boldsymbol{z}') + \frac{1}{C}I\right)\boldsymbol{\alpha} + \mathbf{1}'\boldsymbol{\alpha} \quad (6)$$

where $\boldsymbol{\alpha} = [\alpha_1, \cdots, \alpha_{|\mathcal{I}|}]'$ is the vector of dual variables for the inequality constraints in (3), $\mathcal{A} = \{\boldsymbol{\alpha} | \boldsymbol{\alpha} \geq \mathbf{0}\}$ is the domain of $\boldsymbol{\alpha}$, and $K \in \mathbb{R}^{|\mathcal{I}| \times |\mathcal{I}|}$ is the kernel matrix with each element being $k(x_i, x_j) = \phi(x_i)'\phi(x_j)$.

Next, we show how to solve the min-max optimization problem (6).

### 3.2 Optimization Procedure

#### 3.2.1 Convex Relaxation

The challenge of solving the mixed integer programming (MIP) problem (6) lies in integer programming part $\min_{\boldsymbol{z}}$, which is difficult to solve even for the medium-size integer variables. Existing methods for this MIP problem include semi-definite programming method and alternating optimization method used in transductive SVM (Chapelle et al., 2008). We present a convex relaxation in the spirit of (Li et al., 2009) to problem (6), such that its resultant model is computationally tractable. Specifically, after interchanging the $\max_{\boldsymbol{\alpha} \in \mathcal{A}}$ and $\min_{\boldsymbol{z} \in \mathcal{Z}}$, the optimization problem (6) becomes

$$\max_{\boldsymbol{\alpha} \in \mathcal{A}} \min_{\boldsymbol{z} \in \mathcal{Z}} \mathcal{J}(\boldsymbol{\alpha}, \boldsymbol{z}) \\ = -\frac{1}{2}\boldsymbol{\alpha}'\left(K \odot (\boldsymbol{z}\boldsymbol{z}') + \frac{1}{C}I\right)\boldsymbol{\alpha} + \mathbf{1}'\boldsymbol{\alpha} \quad (7)$$

which is a lower bound of the (6) as (Kim and Boyd, 2008) pointed in the min-max theorem. We introduce another variable $\theta \in \mathbb{R}$ and rewrite (7) as the following quadratically constrained quadratic programming (QCQP) problem

$$\max_{\boldsymbol{\alpha} \in \mathcal{A}, \; \theta \in \mathbb{R}} \; -\theta \\ \text{s.t.} \; \theta \geq -\mathcal{J}(\boldsymbol{\alpha}, \boldsymbol{z}^k), \; \forall \boldsymbol{z}^k \in \mathcal{Z} \quad (8)$$

where $\boldsymbol{z}^k$ is one of $2^{(c^*-1)} - 1$ possible values satisfying the constraints in $\mathcal{Z}$. By introducing $\mu_k$ for each constraint in (8), the partial Lagrangian form of (8) is given by:

$$L(\theta, \boldsymbol{\mu}) = -\theta + \sum_{k: \boldsymbol{z}^k \in \mathcal{Z}} \mu_k(\theta + \mathcal{J}(\boldsymbol{\alpha}, \boldsymbol{z}^k)).$$

where $\boldsymbol{\mu} = \{\mu_k | \forall k : \boldsymbol{z}^k \in \mathcal{Z}\}$. By setting $\frac{\partial L}{\partial \theta} = 0$, we get $\sum \mu_k = 1$. Let $\mathcal{N} = \{\boldsymbol{\mu} | \sum_k \mu_k = 1, \; \mu_k \geq 0\}$ be the domain of $\boldsymbol{\mu}$. By putting $\sum_k \mu_k = 1$ into $L(\theta, \boldsymbol{\mu})$, we obtain the dual form of (8) as

$$\max_{\boldsymbol{\alpha} \in \mathcal{A}} \min_{\boldsymbol{\mu} \in \mathcal{N}} \left\{ \sum_{k: \boldsymbol{z}^k \in \mathcal{Z}} \mu_k \mathcal{J}(\boldsymbol{\alpha}, \boldsymbol{z}^k) = \\ -\frac{1}{2}\boldsymbol{\alpha}'\left(\sum_{k: \boldsymbol{z}^k \in \mathcal{Z}} \mu_k K \odot \boldsymbol{z}^k \boldsymbol{z}^{k'} + \frac{I}{C}\right)\boldsymbol{\alpha} + \mathbf{1}'\boldsymbol{\alpha} \right\} \quad (9)$$

where the equality holds since the objective function is convex in $\boldsymbol{\mu}$ and concave in $\boldsymbol{\alpha}$. With the above convex relaxation, the integer programming $\min_{\boldsymbol{z}}$ in MIP problem (6) is approximated by a convex problem $\min_{\boldsymbol{\mu}}$ in (9) with $\boldsymbol{\mu}$ being a set of continuous variables.

It is also interesting to see that $\max_{\boldsymbol{\alpha}} \min_{\boldsymbol{\mu}}$ in (9) shares the similar forms with multiple kernel learning (MKL) (Lanckriet et al., 2004; Rakotomamonjy et al., 2008). More exactly, taking $\boldsymbol{\mu}$ as the set of mixing coefficients, term $\sum_{k:\boldsymbol{z}^k \in \mathcal{Z}} \mu_k K \odot \boldsymbol{z}^k \boldsymbol{z}^{k'}$ can be seen as the convex combination of $|\mathcal{Z}|$ base kernel matrices $K \odot \boldsymbol{z}^k \boldsymbol{z}^{k'}$, and each base kernel is constructed from a feasible label vector $\boldsymbol{z}^k \in \mathcal{Z}$. Therefore, we can use MKL solvers to solve problem (9).

#### 3.2.2 Cutting Plane Method

Since the number of constraints in (8) (i.e., the size of the set of base kernels in (9)) is $2^{(c^*-1)} - 1$, it is still computationally expensive to solve (9) for large $c^*$. Considering not all constraints in (8) are active at optimality, the cutting plane algorithm (Kelley, 1960) can be used to iteratively select a small set of constraints, denoted by $\mathcal{Z}_c \subset \mathcal{Z}$, to solve MKL problem (9). The procedure is as follows: 1) initialize $\mathcal{Z}_c$ by the most violated variable $\boldsymbol{z}_v \in \mathcal{Z}$; 2) solve MKL problem (9) using the constraints in $\mathcal{Z}_c$ instead of $\mathcal{Z}$; 3) find the next most violated $\boldsymbol{z}_v$ from $\mathcal{Z}$ and add it in $\mathcal{Z}_c$. Step 2 and 3 are iteratively conducted until convergence.

Finding the most violated constraint $\boldsymbol{z}_v$ from $\mathcal{Z}$ is achieved by solving the optimization problem: $\max_{\boldsymbol{z} \in \mathcal{Z}} \left\| \sum_{j \in \mathcal{I}} \alpha_j z_j \varphi(x_j) \right\|$. By replacing this $\ell_2$ norm with $\ell_\infty$ norm and letting [1] $\varphi(x_j) = [x_{j1}, x_{j2}, \cdots, x_{js}]$ where $s$ is the dimension of $\varphi(x_j)$, this optimization is same as: $\max_{\iota=1,\cdots,s} \left( \max_{\boldsymbol{z} \in \mathcal{Z}} \left| \sum_{j \in \mathcal{I}} \alpha_j x_{j\iota} z_j \right| \right)$. By analyzing the quantities of coefficient scalers $\max_\iota \alpha_j x_{j\iota}$, this linear integer programming problem can be easily solved without resorting to any numerical optimization solver. Note that due to Remark 1 only $c^*$ elements in $\boldsymbol{z}_v$ need to be determined.

#### 3.2.3 Multiple Kernel Learning

In this paper, the problem (9) with the constraint set $\mathcal{Z}_c$ is solved by the SimpleMKL method (Rakotomamonjy et al., 2008), in which an alternating method is used. First, when $\boldsymbol{\mu}$ is fixed, one needs to solve the SVM dual as follows

---
[1] if $\varphi(x_j)$ has infinite dimensions, we perform singular value decomposition for kernel matrix to get $\varphi(x_j) = [x_{j1}, x_{j2}, \cdots, x_{js}]$.

$$\max_{\boldsymbol{\alpha}\in\mathcal{A}} -\frac{1}{2}\boldsymbol{\alpha}'\left(\sum_{k:\boldsymbol{z}^k\in\mathcal{Z}_c}\mu_k K\odot \boldsymbol{z}^k\boldsymbol{z}^{k'}+\frac{1}{C}\Lambda\right)\boldsymbol{\alpha}+\mathbf{1}'\boldsymbol{\alpha}.$$

Then, when $\boldsymbol{\alpha}$ is fixed, the reduced gradient algorithm can be used to update $\boldsymbol{\mu}$. These two procedures are iteratively conducted until convergence.

### 3.2.4 Determining the Optimal $z$

The optimal $\boldsymbol{z}$ for the model (3) is determined based on the results of the MKL problem (9), i.e., $\mathcal{Z}_c$ and $\boldsymbol{\mu}$. Due to Remark 1, each vector $\boldsymbol{z}^k \in \mathcal{Z}_c$ with $|\mathcal{I}|$ elements can be compressed as a vector $\boldsymbol{z}^k_{c^*}$ with $c^*$ elements. Each element of $[\boldsymbol{z}^k_{c^*}]_i$, $i=1,\cdots,c^*$ is same as any label $z$ in $\boldsymbol{z}^k$ for data in class $\omega_i$. For the same reason, we only need to determine the compressed vector $\boldsymbol{z}_{c^*}$ with $c^*$ elements instead of $\boldsymbol{z}$ with $|\mathcal{I}|$ elements. In the proposed method, we determine $\boldsymbol{z}_{c^*}$ by using a graph cut method on the weighted label matrix $A=\sum_{k:\boldsymbol{z}^k\in\mathcal{Z}_c}\mu_k\boldsymbol{z}^k_{c^*}\boldsymbol{z}^{k'}_{c^*}\in R^{c^*\times c^*}$, which represents the correlations among labels. For simplicity, maximum spanning tree algorithm, i.e. Kruskal's algorithm, is adopted as the graph cut method.

### 3.3 Overall Scheme of Generating Class Structure Tree

Now, we proceed to present the overall scheme for generating the whole class structure tree. As shown in Algorithm 1, all nodes in the tree are determined from top to bottom and from left to right. Specifically, at Steps 1 and 2, the root node $G_1^1$, the layer index $t$ and node index $i$ are initialized. At Steps 3 and 4, $\boldsymbol{\alpha}$ and $\mathcal{Z}_c=\{\boldsymbol{z}\}$ are initialized to be $\boldsymbol{\alpha}=\frac{1}{|\mathcal{I}|}\mathbf{1}$ and $\mathcal{Z}_c=\{\boldsymbol{z}\}$. At Steps 5 and 6, the cutting plane algorithm and MKL algorithm are iteratively solved to get the set of multiple labels $\mathcal{Z}_c$ and the set of corresponding weight parameters $\boldsymbol{\mu}$. The convergence of the MKL procedure has been discussed in (Kim and Boyd, 2008; Rakotomamonjy et al., 2008). At Steps 7 and 8, the optimal $\boldsymbol{z}$ and two child nodes are determined. At Step 9, the next node selected and the algorithm goes to Step 3 again. The procedure (Steps 3-9) is repeated until all leaf nodes are reached.

**Remark 2** *In Step 5 of Algorithm 1, running MKL requires the pre-stored kernel matrix which could be huge for large datasets. However, when linear (or polynomial) kernel is used, storing kernel matrix can be avoided and MKL can be efficiently solved by calling fast linear SVM solvers e.g. (Hsieh et al., 2008; Joachims, 2006). Specifically, following (Rakotomamonjy et al., 2008), MKL problem (9) can be equiva-lently written as*

$$\min_{\boldsymbol{\mu}\in\mathcal{N},\{w_k\},\boldsymbol{\xi}\geq\mathbf{0}} \sum_{k:\boldsymbol{z}^k\in\mathcal{Z}}\|w_k\|^2+C\sum_{j\in\mathcal{I}}\xi_j^2 \\ s.t \quad \sum_k z_j^k\mu_k w_k'x_j\geq 1-\xi_j, j\in\mathcal{I}. \quad (10)$$

*By introducing augmented variables $\hat{w}=[w_1',\cdots,w_{|\mathcal{Z}|}']'$ and $\psi(x_j,\boldsymbol{\mu})=[z_j^1\mu_1 x_j',\cdots,z_j^{|\mathcal{Z}|}\mu_{|\mathcal{Z}|}x_j']'$, problem (10) can be rewritten as*

$$\min_{\boldsymbol{\mu}\in\mathcal{N},\hat{w},\boldsymbol{\xi}\geq\mathbf{0}} \|\hat{w}\|^2+C\sum_{j\in\mathcal{I}}\xi_j^2 \\ s.t \quad \hat{w}'\psi(x_j,\boldsymbol{\mu})\geq 1-\xi_j, j\in\mathcal{I}. \quad (11)$$

*When $\boldsymbol{\mu}$ is fixed, model (11) has the same manner as linear SVM model. Therefore, existing solvers (Hsieh et al., 2008; Joachims, 2006) are applicable.*

---

**Algorithm 1** The Algorithm of MSM

**Input:** Data $\mathcal{D}$.
**Output:** The class structure tree.
1. Initialize the root node to be $\{1,\cdots,c\}$.
2. Let $t=1$ and $i=1$.
**repeat**
   3. For non-leaf node $G_i^t$, initialize $\boldsymbol{\alpha}=\frac{1}{|\mathcal{I}|}\mathbf{1}$.
   4. Find the most violated $\boldsymbol{z}\in\mathcal{Z}$ and let $\mathcal{Z}_c=\{\boldsymbol{z}\}$.
   **repeat**
     5. Run MKL algorithm with constraint set $\mathcal{Z}_c$.
     6. Find the most violated $\boldsymbol{z}$ and set $\mathcal{Z}_c=\mathcal{Z}_c\cup\boldsymbol{z}$.
   **until** Steps 5-6 converge
   7. Determine optimal $\boldsymbol{z}$ by graph cut algorithm.
   8. Generate child nodes by splitting rule (4).
   9. Find the next non-leaf node and update $t,i$.
**until** All leaf nodes are reached

Table 1: Characteristics of datasets used in the experiments. Each dataset is split into training dataset $\mathcal{D}_{trn}$ and testing dataset $\mathcal{D}_{tst}$. $c$ is number of classes and $s$ is the number of features.

| Dataset | $|\mathcal{D}_{trn}|$ | $|\mathcal{D}_{tst}|$ | $c$ | $s$ |
|---|---|---|---|---|
| svmguide4 | 300 | 312 | 6 | 10 |
| vowel | 528 | 462 | 11 | 10 |
| segment | 1000 | 1310 | 7 | 19 |
| satimage | 2000 | 4435 | 6 | 36 |
| usps | 2007 | 7291 | 10 | 256 |
| news20 | 3993 | 15935 | 20 | 62061 |
| sector | 3207 | 6412 | 105 | 55197 |
| Caltech 101 | 3030 | 5647 | 101 | 1000 |

## 4 Numerical Experiments

### 4.1 Datasets

In experiments, we test the proposed method and benchmark methods on eight datasets and their characteristics are given in Table 1. Among them,

Table 2: Mean and standard derivation of testing accuracy (in %) of seven methods. Best performance among all methods is highlighted in bold.

| Kernel | Dataset | MSM | I-BWG | J-BWG | FT | CPT | 1vs1 | 1vsR |
|---|---|---|---|---|---|---|---|---|
| Gaussian | svmguide4 | 77.51±0.42 | 72.77±0.81 | — | 60.70±2.32 | 72.41±0.23 | **80.45±0.51** | 77.13±0.42 |
| | vowel | **91.42±0.33** | 70.89±1.52 | — | 70.76±3.56 | 76.58±0.45 | 87.98±0.33 | 83.67±0.21 |
| | segment | **92.57±0.22** | 89.56±1.44 | — | 89.20±1.44 | 90.48±0.10 | 92.31±0.42 | 91.29±0.43 |
| | satimage | **87.45±0.61** | 72.29±1.10 | — | 60.07±2.89 | 69.64±0.45 | 61.75±0.23 | 81.99±0.33 |
| | usps | **94.31±0.32** | 69.98±0.34 | — | 78.94±0.98 | 84.03±4.81 | 87.99±0.21 | 90.29±0.24 |
| Linear | svmguide4 | 70.84±1.16 | 65.17±1.04 | **86.67±1.05** | 68.82±2.21 | 53.01±0.79 | 81.71±0.96 | 70.83±1.07 |
| | vowel | 57.74±2.25 | 35.81±1.75 | 50.55±1.46 | 30.18±3.54 | 42.52±0.88 | **67.86±1.16** | 38.31±2.56 |
| | segment | **93.22±0.08** | 69.04±0.04 | 83.18±0.06 | 74.31±4.87 | 84.25±0.06 | 92.14±0.05 | 92.13±0.02 |
| | satimage | **78.81±0.14** | 68.22±0.10 | 70.34±0.11 | 63.90±2.26 | 68.32±0.15 | 77.78±0.09 | 75.42±0.09 |
| | usps | 84.30±0.07 | 78.21±0.11 | 78.75±0.11 | 57.08±2.10 | 85.95±0.09 | **93.87±0.02** | 91.61±0.06 |
| | news20 | **82.31±0.10** | 70.09±0.12 | 70.18±0.10 | 42.29±4.26 | 65.63±0.11 | 77.52±0.07 | 78.65±0.08 |
| | sector | **88.01±0.08** | 80.17±0.07 | 80.90±0.09 | 79.90±5.56 | 73.61±0.10 | 86.14±0.05 | **88.01±0.06** |
| | Caltech 101 | **39.78±0.03** | 13.84±0.32 | 13.31±0.20 | 16.33±0.30 | 24.53±0.09 | 36.64±0.10 | 39.50± 0.20 |

`svmguide4, vowel, segment, satimage, usps, sector` and `news20` are downloadable from LIBSVM website [2], and `Caltech 101` is provided by the archive of computational vision at Caltech [3].

For dataset `Caltech 101`, we follow (Lazebnik et al., 2006) to exclude the background class in which none of the images belong to the defined categories. We randomly select 30 images from each class for training and test on the remainder. As usual, SIFT feature (Lowe, 2004) is adopted due to its good performance. Specifically, we use a dense grid sampling strategy to select regions of interest, and the step size and patch size are fixed to 8 and 16 respectively. The maximum side (width/length) of each image is resized to 300 pixels. The codebook size is fixed to be 1000, and $k$-means is simply used on all features to generate the codebook.

### 4.2 Experimental Setting

The proposed method MSM is compared with two traditional methods 1vs1 and 1vsR as well as three recent class structure learning method FT (Beygelzimer et al., 2009b), CPT (Beygelzimer et al., 2009a) and BWG (Bengio et al., 2010). In BWG, two implementations are provided: learn each tree node individually or learn all nodes jointly. In the sequel, we term these two implementations as I-BWG and J-BWG respectively. For the sake of fair comparison, I-BWG and J-BWG are restricted to use binary tree in all experiments.

The experiments are done on a linux machine with 2.27GHZ Intel(R) Core(TM) 4 DUO CPU and 24GB memory. Gaussian kernel $k(x_i, x_j) = \exp(-\eta\|x_i - x_j\|^2)$ and linear kernel $k(x_i, x_j) = x_i' x_j$ are both used in experiments. However, considering the computational cost, Gaussian kernel is neither used for large datasets nor used in the method J-BWG. In all methods, the SVM problem with Gaussian and linear kernel are solved by sequential minimal optimization algorithm (Platt, 1998) and dual coordinate descent algorithm (Hsieh et al., 2008), respectively. The regularization parameter $C$ in SVM and the Gaussian kernel parameter $\eta$ are chosen by cross-validation procedure among the range [0.001, 0.01, 0.1, 1, 10, 100, 1000]. With the tuned parameters, all methods are run for 7 realizations, each of which has different random splitting of $\mathcal{D}_{trn}$ and $\mathcal{D}_{tst}$ with fixed ratio $\frac{|\mathcal{D}_{trn}|}{|\mathcal{D}_{tst}|}$ as given in Table 1.

To simulate the scenario of instantaneous prediction, each method predicts one testing instance's label by traversing all needed binary classifiers once. The timing and accuracy of testing procedure are compared for all methods. Here, the prediction accuracy is defined as the mean accuracy among all classes, i.e., the average of the diagonal vector of confusion matrix.

### 4.3 Experimental Results

To illustrate the learned class structure, Figure 1 shows the class structure tree learned by MSM, I-BWG, FT and CPT with Gaussian kernel for usps dataset. The usps dataset contains images of 10 handwritten digits "0 ∼ 9". As seen from the original images in the website http://www.cs.nyu.edu/∼roweis/data.html, some pairs of digits are endowed with similar patterns, e.g., pairs "1-7", "4-6", "8-9", "9-0" and "8-0". Therefore, it is relatively harder to discriminate two digits for these confusing pairs. From the figures, MSM can group those alike (confusing) digits into one category even at the next layer to the bottom, which is consistent with the motivation of MSM method. I-BWG method succeeds to group some alike (confusing) digits together in the second layer but fails in the deeper layers. FT method completely fails due to its randomness. CPT performs better than I-BWG and FT but fails to group "1-7" together in the third layer.

Table 2 shows the mean and standard derivation of the accuracies of seven methods on all datasets for both Gaussian and linear kernels. From this table, the proposed method MSM can achieve the best performance among all seven methods for 9 of 13 dataset settings. For the rest of 4 dataset settings, the best performances are achieved by methods 1vs1, 1vsR and J-BWG, but MSM is still able to be ranked as top 2 or 3 among seven methods. The better performance of MSM over I-BWG and J-BWG is probably due to the fact that the confusion matrix used in both I-BWG and J-BWG heavily depends the results of 1vsR model, as shown in Table 2, which may not be accurate and cannot effectively reflect the similarity among classes. Thus, their class structure may not be as good as that of MSM, which directly uses maximum separating margin criterion for tree construction. FT does not yield good results, since it has theoretical pitfall, and the advantage of MSM over CPT is plausibly attributed to that all training instances are considered together in MSM. MSM also outperforms 1vs1 and 1vsR models, which demonstrates that the learned class structure of MSM helps to improve the prediction for multi-class problems.

---

[2] http://www.csie.ntu.edu.tw/∼cjlin/libsvmtools/datasets/
[3] http://www.vision.caltech.edu/Image_Datasets/

Table 3 shows the prediction timing results of seven methods on all datasets for both kernels. As seen from this table, five class structure learning methods spend nearly comparable time on prediction for all dataset settings, and these methods are $2 \sim 200$ times faster than 1vs1 and $1.3 \sim 50$ times faster than 1vsR. Figure 2 plots curves of prediction timing of all methods against $c$ for news20 dataset. From this figure, the slopes for the curves of 1vs1 and 1vsR are larger than those of class structure learning methods for all $c$. This observation is consistent with their testing computational complexities as mentioned before.

## 5 Conclusion

This paper presents a novel method to determine the class structure tree for multi-class classification. First, we propose a separating margin criterion to measure the goodness of the child nodes for any internal node of the tree. Then, with this criterion, we propose a maximum separating margin model to determine the child nodes for any internal node. The proposed model is efficiently solved by using a mild convex relaxation method. The effectiveness and efficiency of the proposed method are evidently shown in real world large-scale datasets.

### Acknowledgements

This research was in part supported by Singapore MOE AcRF Tier-1 Research Grant (RG15/08) and A* SERC Grant (102 158 0034).

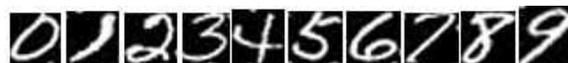

(a)

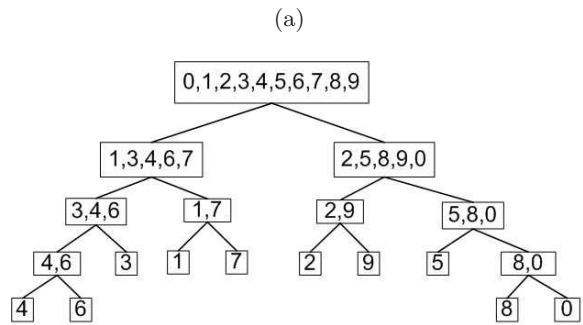

(b) MSM

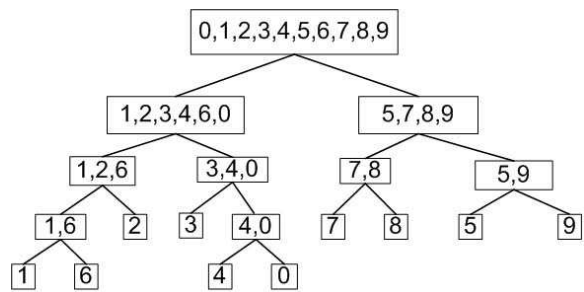

(c) I-BWG

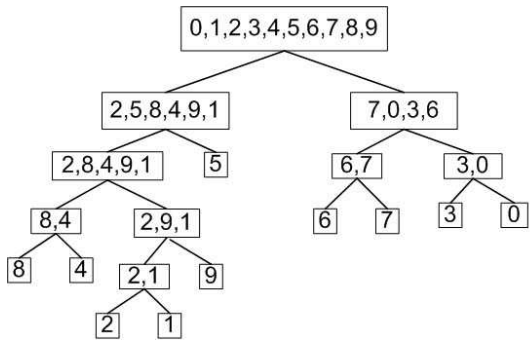

(d) FT

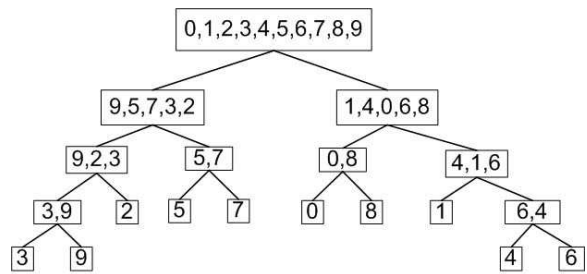

(e) CPT

Figure 1: Figure (a) is 10 samples of images for ten digits "$0 \sim 9$" in usps dataset. Figures (b) to (e) are class structure trees generated by methods MSM, I-BWG, FT and CPT, with Gaussian kernel, respectively.

Table 3: Prediction time of seven methods (in second).

| Kernel | Dataset | MSM | I-BWG | J-BWG | FT | CPT | 1vs1 | 1vsR |
|---|---|---|---|---|---|---|---|---|
| Gaussian | svmguide4 | 0.03 | 0.06 | — | 0.04 | 0.04 | 0.22 | 0.08 |
| | vowel | 0.06 | 0.09 | — | 0.08 | 0.08 | 1.45 | 0.22 |
| | segment | 0.19 | 0.17 | — | 0.20 | 0.18 | 1.78 | 0.59 |
| | satimage | 1.59 | 1.79 | — | 1.63 | 1.98 | 9.66 | 4.22 |
| | usps | 27.49 | 26.5 | — | 28.33 | 27.96 | 205.24 | 56.94 |
| Linear | svmguide4 | 0.03 | 0.03 | 0.02 | 0.03 | 0.03 | 0.03 | 0.13 |
| | vowel | 0.05 | 0.05 | 0.05 | 0.05 | 0.05 | 0.09 | 0.31 |
| | segment | 0.13 | 0.13 | 0.13 | 0.13 | 0.13 | 0.13 | 0.41 |
| | satimage | 0.35 | 0.36 | 0.38 | 0.36 | 0.38 | 0.34 | 0.83 |
| | usps | 0.87 | 0.88 | 0.88 | 0.87 | 0.88 | 2.61 | 1.70 |
| | news20 | 754.27 | 647.00 | 546.18 | 764.25 | 758.11 | 14914.00 | 1635.41 |
| | sector | 291.31 | 285.52 | 201.28 | 288.22 | 271.75 | 181326.21 | 5060.37 |
| | Caltech 101 | 5.91 | 4.33 | 4.81 | 5.72 | 5.03 | 886.41 | 33.58 |

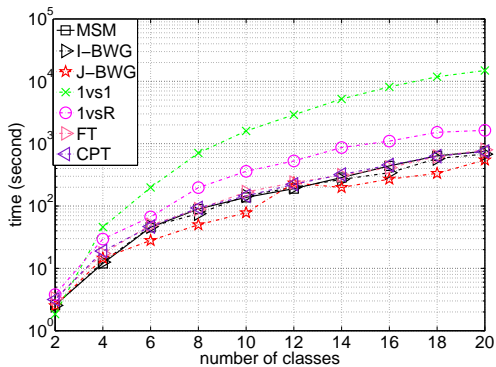

Figure 2: The plots of prediction time against the number of classes for news20 dataset. The classes are randomly selected among 20 classes.